\documentclass[conference]{IEEEtran}

\usepackage{cite}

\usepackage{graphicx}
\usepackage{subcaption}
\usepackage[pagebackref=true,breaklinks=true,letterpaper=true,colorlinks,bookmarks=false]{hyperref}

\usepackage{times}
\usepackage{epsfig}
\usepackage{amsmath}
\usepackage{amssymb}
\usepackage{multirow}
\usepackage{lipsum}
\usepackage{acronym}

\newcommand*{\ie}		{i.e.,\ }
\newcommand*{\etal}		{et~al.}
\newcommand*{\sota}		{state-of-the-art}

\acrodef{cnn}		[\textsc{CNN}]				{Convolutional Neural Network}
\acrodef{fcnn}		[\textsc{FCNN}]				{Fully Convolutional Neural Network}
\acrodef{iou}		[\textsc{IoU}]				{Intersection over Union}
\acrodef{ilsvrc}	[\textsc{ILSVRC}]			{ImageNet Large Scale Visual Recognition Competition}
\acrodef{gt}	    [\textsc{GT}]   			{ground truth}
\acrodef{crf}	    [\textsc{CRF}]   			{Conditional Random Field}
\acrodef{pspnet}	[\textsc{PSPNet}] 			{Pyramid Scene Parsing Network}
\acrodef{voc}       [\textsc{VOC}]              {Visual Object Classes}

\def\footnoterule{\relax%
  \kern-5pt
  \hbox to \columnwidth{\hfill\vrule width \columnwidth height 0.4pt\hfill}
  \kern4.6pt}

\begin{document}

\title{Recognizing Challenging Handwritten Annotations with Fully Convolutional Networks}


\author{
    \IEEEauthorblockN{Andreas K\"olsch\IEEEauthorrefmark{1}\IEEEauthorrefmark{2},
    Ashutosh Mishra\IEEEauthorrefmark{1},
    Saurabh Varshneya\IEEEauthorrefmark{1}\IEEEauthorrefmark{2},
    Muhammad Zeshan Afzal\IEEEauthorrefmark{1}\IEEEauthorrefmark{2},
    Marcus Liwicki\IEEEauthorrefmark{1}\IEEEauthorrefmark{2}\IEEEauthorrefmark{3}\IEEEauthorrefmark{4}}
    
    \IEEEauthorblockA{
            \{a\_koelsch12, a\_ashutosh16, s\_varshney16\}@cs.uni-kl.de, afzal@iupr.com, marcus.liwicki@ltu.se
    }
    
    \vspace{\baselineskip}
    
    \IEEEauthorblockA{\IEEEauthorrefmark{1}MindGarage, University of Kaiserslautern, Germany}
    \IEEEauthorblockA{\IEEEauthorrefmark{2}Insiders Technologies GmbH, Kaiserslautern, Germany}
    \IEEEauthorblockA{\IEEEauthorrefmark{3}University of Fribourg, Switzerland}
    \IEEEauthorblockA{\IEEEauthorrefmark{4}Lule\aa,  University of Technology, Sweden}
}

\maketitle

\begin{abstract}
This paper introduces a very challenging dataset of historic German documents and evaluates \ac{fcnn} based methods to locate handwritten annotations of any kind in these documents. The handwritten annotations can appear in form of underlines and text by using various writing instruments, e.g., the use of pencils makes the data more challenging. We train and evaluate various end-to-end semantic segmentation approaches and report the results. The task is to classify the pixels of documents into two classes: background and hand\-written annotation. The best model achieves a mean \ac{iou} score of $95.6\,\%$ on the test documents of the presented dataset. We also present a comparison of different strategies used for data augmentation and training on our presented dataset. For evaluation, we use the Layout Analysis Evaluator for the ICDAR 2017 Competition on Layout Analysis for Challenging Medieval Manuscripts.

\end{abstract}

\section{Introduction}
Libraries are often interested in analyzing handwritten annotations in historic manuscripts or prints.
Such annotations can give hints to the provenience of the documents or provide additional information about the readers' thoughts.

For successful handwriting recognition in document images, a multi-stage processing pipeline is needed.
Before an algorithm can analyze handwritten annotations in a document, it is important to localize the areas in this document that contain handwriting.
Therefore, it is crucial to implement a robust algorithm for finding these areas as one of the first steps in the processing pipeline.

Due to the huge variety in the layout of these documents, it is very difficult to come up with a rule-based algorithm that can reliably find handwritten annotations in unseen documents. Many previous approaches aim to segment the documents into regions in a top-down manner, bottom-up or, using texture-based methods~\cite{asi2015simplifying}.

In the last years, deep learning based methods have been very successful in many computer vision tasks, such as classification~\cite{russakovsky2015imagenet,simonyan2014very,Szegedy_2015_CVPR,DBLP:journals/corr/HeZRS15} and segmentation of real-world images~\cite{everingham2015pascal,Long_2015_CVPR,chen2018deeplab}. 
There are two distinct approaches for semantic segmentation using different types of network architectures. These approaches are based on \acfp{cnn}~\cite{everingham2015pascal,Long_2015_CVPR,chen2018deeplab} and LSTMs (Long Short-Term Memory) neural networks~\cite{graves2012supervised,Afzal_sltm_binarization}. Although the results are promising with both the approaches, the \ac{cnn} based approaches are becoming popular because of their efficiency. These networks which are known as \acfp{fcnn} outperformed previous methods.
Most recently, \acp{fcnn} have also been used to segment historical document images~\cite{simistira2017icdar2017,chen2017convolutional,xu2017page}. The methods we evaluate on our dataset are \ac{fcnn} based.

\begin{figure}
    \centering
        \fbox{\includegraphics[width=0.46\linewidth]{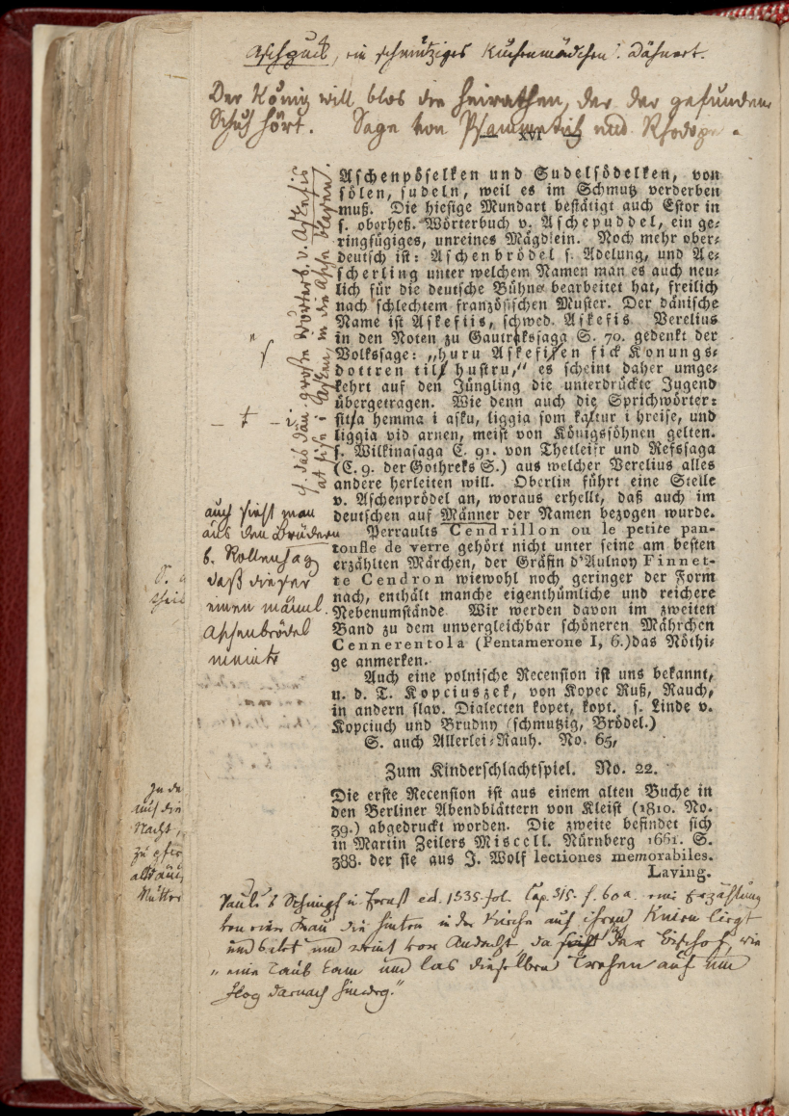}}
        \fbox{\includegraphics[width=0.46\linewidth]{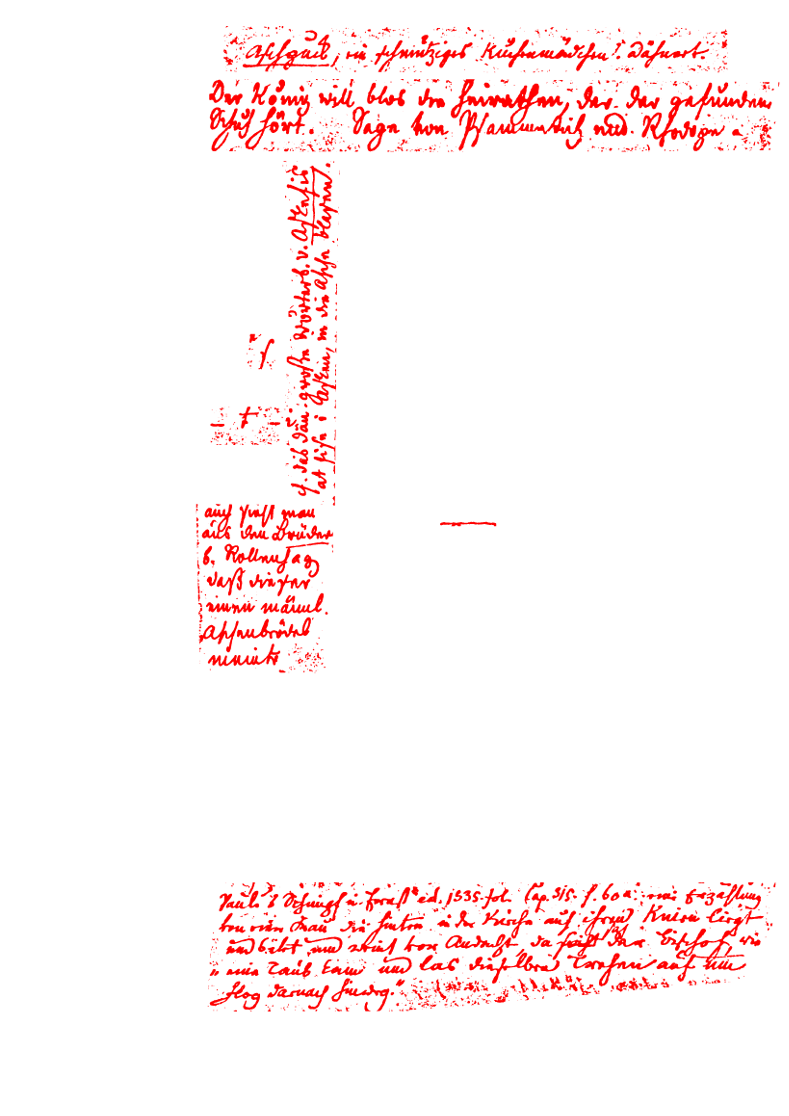}}
    \caption[]{Sample image from the presented dataset. The left image is the original image\footnotemark, the right image shows the \ac{gt}.}
    \label{fig:intro_dataset}
\end{figure}

\begin{figure*}
    \centering
        \fbox{\includegraphics[width=0.45\linewidth,height=0.15\linewidth]{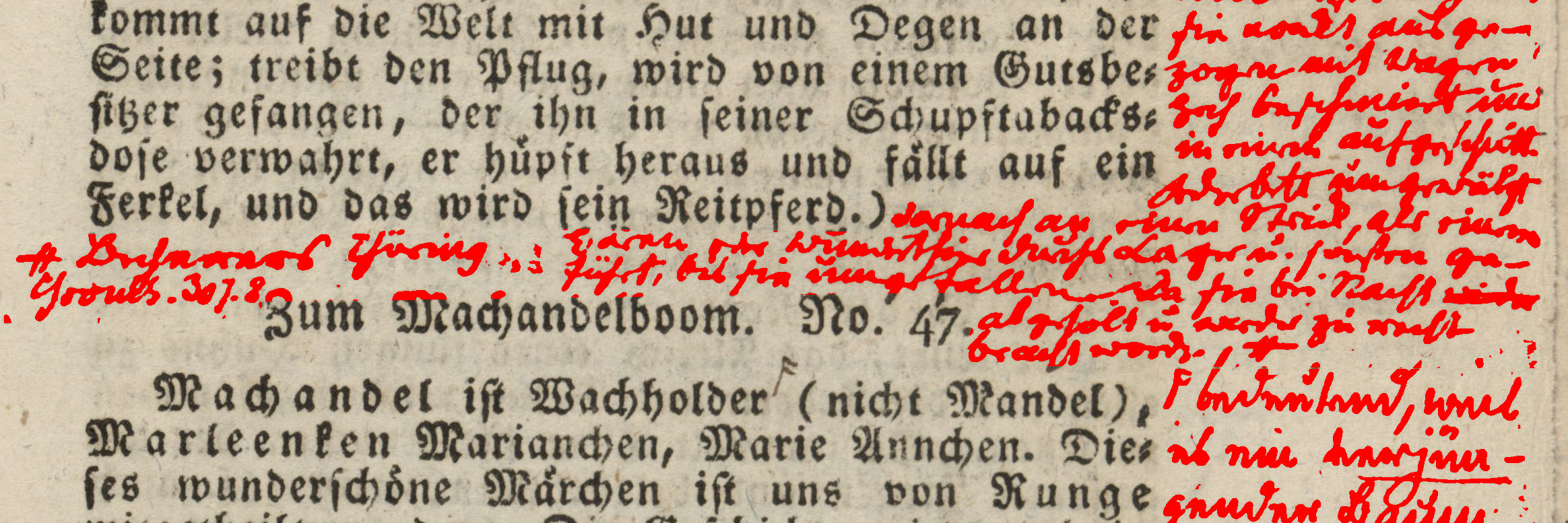}}
        \fbox{\includegraphics[width=0.45\linewidth,height=0.15\linewidth]{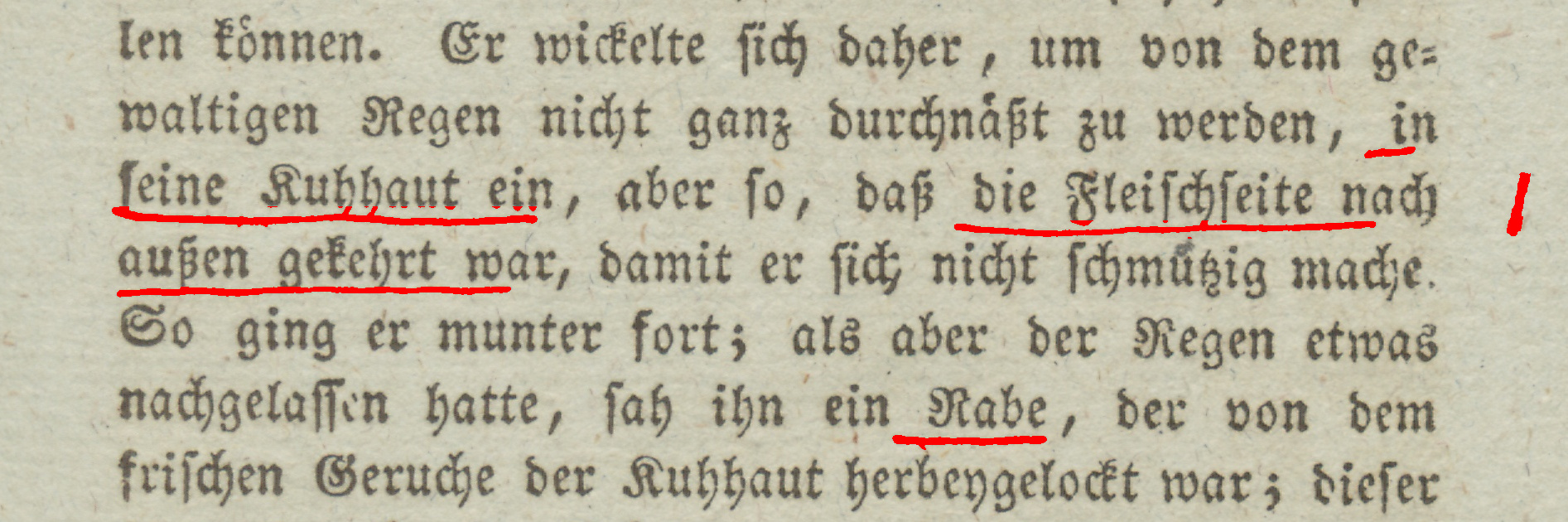}}
        \par\smallskip
        \fbox{\includegraphics[width=0.45\linewidth,height=0.15\linewidth]{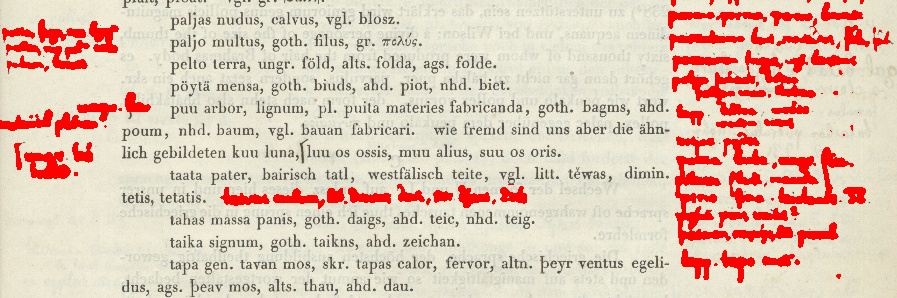}}
        \fbox{\includegraphics[width=0.45\linewidth,height=0.15\linewidth]{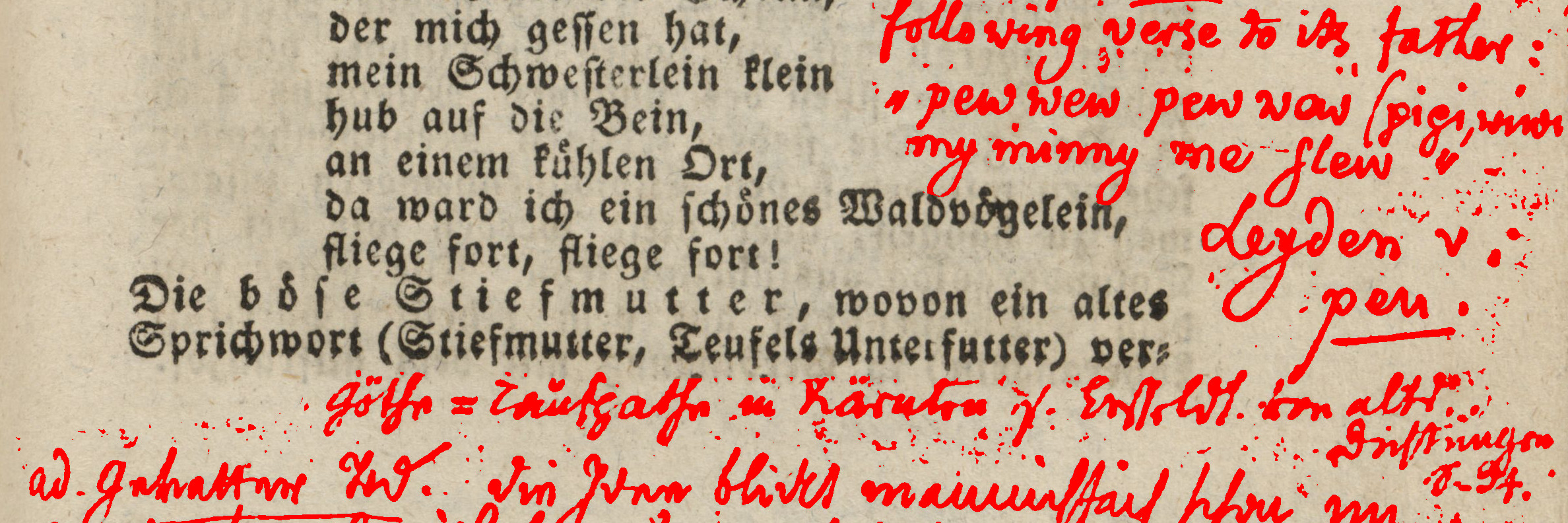}}
        \par\smallskip
        \fbox{\includegraphics[width=0.45\linewidth,height=0.15\linewidth]{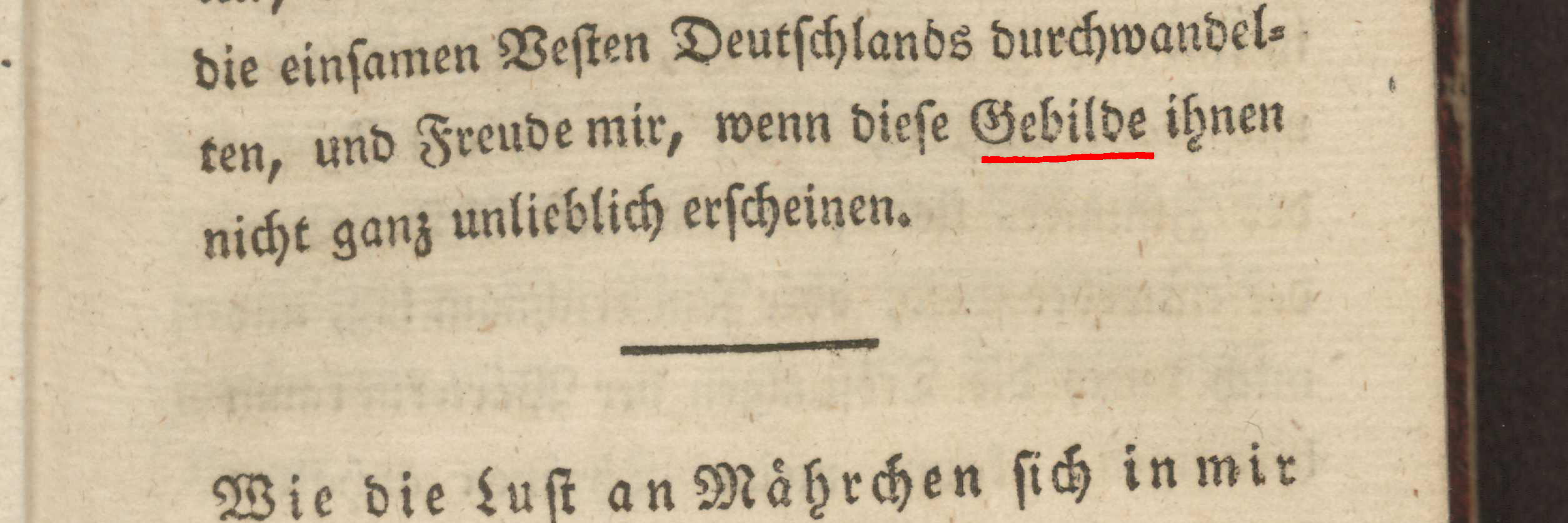}}
        \fbox{\includegraphics[width=0.45\linewidth,height=0.15\linewidth]{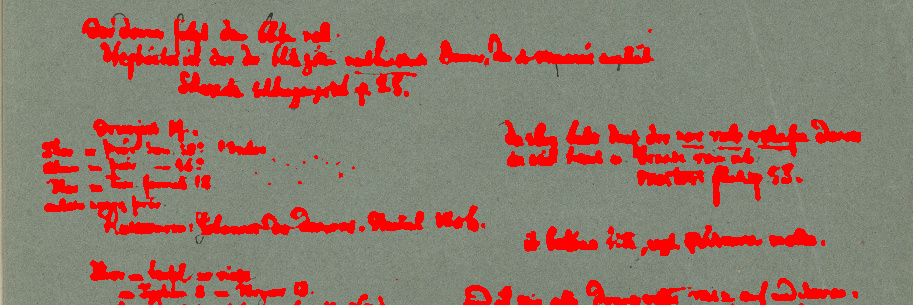}}
    \caption{Sample patches of the document images with the \ac{gt} as an overlay (red) depicting the different challenges of the dataset. Image Sources are described in Section~\ref{sec:dataset}.}
    \label{fig:dataset}
\end{figure*}

Our dataset consists of $50$ images which are divided into $40$ representative training images and $10$ test images. The images are taken from multiple manuscripts and feature different layouts and various kinds of annotations (cf.~Section~\ref{sec:dataset}).
For each image, there exists a pixel-wise \acf{gt} with two classes: handwritten annotation and background (cf.~Fig.~\ref{fig:intro_dataset}).
The task associated with this dataset is to classify each pixel of the input images and to maximize the mean \ac{iou} score.

\footnotetext{Universit\"atsbibliothek der Humboldt-Universit\"at zu Berlin, Signatur 0001 1433243313511 1 221, Seite 1}

The \textbf{contribution} of our work is twofold: First, we present a new challenging dataset for handwritten annotation detection. Secondly, we train and evaluate multiple methods for segmentation on this dataset.

The remaining sections of this paper are organized as follows: Section~\ref{sec:related} briefly describes related work in the fields of both handwriting recognition and deep learning. In Section~\ref{sec:dataset}, the new dataset and its unique challenges are illuminated. The network architectures we use are presented in Section~\ref{sec:networks}.
The experimental setup and training details are described in Section~\ref{sec:experiments}. Section~\ref{sec:results} reports the results of the experiments that is followed by the discussion in Section~\ref{sec:discussion}. At the last, Section~\ref{sec:conclusion} concludes the paper and gives perspectives for future work.



\section{Related Work}
\label{sec:related}
Traditional approaches for the semantic segmentation of documents build on machine learning methods applied to hand-crafted features.
Typically, the contribution is extracting good feature representations of the document and feeding them to a classifier trained on the training data~\cite{ZAGORIS20141051,6977226,chen2015page}.

Zagoris \etal \ used Bag of Visual Words model to extract local image
features which are fed to an SVM based classifier for segmentation of key points~\cite{ZAGORIS20141051}. This approach, however, requires binarization and Hough transform 
as preprocessing operations which can be dependant on the type of the documents.

Chen \etal~\cite{6977226} extract color and coordinate based handcrafted feature representations from the colored images, and the obtained features are fed to an SVM classifier. The method works directly on colored images without any pre-processing operations like binarization. However, a threshold based post-processing step is required which can be dependent on training data.

Deep learning has been the key success factor for many of the fields related to computer vision and pattern recognition~\cite{lecun2015deep} such as document analysis~\cite{lecun2015deep, liwicki2016latest}. Recently, many approaches for document classification~\cite{deepdoc_afzal,afzal2017cutting,kolsch2017real}, binarization~\cite{pastor2015insights,Afzal_sltm_binarization, tensmeyer2017document}, historical document segmentation~\cite{chen2015page,chen2017convolutional,younas2017d}, image format detection~\cite{trier2017deep}, multiple script identification~\cite{ul2015sequence, fujii2017sequence}, character recognition of difficult scripts~\cite{AhmadARLB15, Ahmed2016} etc. Following is the description of closely related methods of semantic segmentation that put the proposed work into perspective.

Very recent approaches for the segmentation of historical documents use features extracted from deep-learning based architectures like auto-encoders. Such architectures learn the features automatically from unlabeled given data. The segmentation task here is modeled as a pixel-labeling task where each pixel is assigned a label in the image.
Chen \etal \cite{chen2015page} used features extracted from auto-encoders which are classified using an SVM classifier.

\ac{cnn} based approaches have also been used in a similar fashion where features are obtained automatically from the last or second last fully connected layer of a \ac{cnn}.
Chen and Seuret \cite{chen2017convolutional} generate small image patches using a superpixel algorithm and then a \ac{cnn} is applied  over the fixed-sized patches to classify image patches called as superpixels into respective classes.

Most similar to our work is the application of \ac{fcnn} based methods which have successfully been applied to various datasets~\cite{Long_2015_CVPR,chen2018deeplab,xu2017page}.
In case of documents, Yang \etal \ use a multimodal \ac{fcnn} for extracting semantic structures from document images~\cite{DBLP:journals/corr/YangYAKKG17}.
In addition to \ac{fcnn}, it combines the unsupervised reconstruction task with the pixel-labeling task and added a text embedding map to take into account the content of underlying text for better extraction of semantic structures like figure, table, heading, and paragraph. These methods, however, have been applied to more homogeneous data.
In this paper, we study the effectiveness of \acp{fcnn} on diverse data of challenging historical documents.

\section{The Dataset}
\label{sec:dataset}

The actual success or failure of any model is truly dependent on the type of dataset used for its training and testing.
In the last decade, rich sets of historical manuscript databases have been collected, such as DIVA-HisDB~\cite{simistira2016diva}, ENP~\cite{7333898} and IMPACT~\cite{Papadopoulos:2013:IDH:2501115.2501130}.
For our experiments, we introduce a challenging dataset for training and benchmarking our approaches for handwritten annotation segmentation.

Our new dataset comprises 40 images for training and validation and 10 images for testing.
The images with their respective \ac{gt} in the well-known PAGE format~\cite{pletschacher2010page} are kindly provided by 
\begin{itemize}
    \item Universit\"atsbibliothek der Humboldt-Universit\"at zu Berlin
    \item Universit\"atsbibliothek Kassel
    \item Staatsbibliothek Berlin 
    \item Staatsarchiv Marburg 
\end{itemize}
Figure~\ref{fig:dataset} illustrates different patches of sample images from our dataset.

An interesting feature of the dataset is that it contains document pages from multiple sources which are digitized using different devices. This increased variance makes our dataset especially challenging for segmentation tasks.
While in other datasets the background and foreground colors of the images are typically similar, they are very heterogeneous in our dataset.
Not only do the font styles and sizes in our dataset vary significantly for both printed text and annotations, but also the line spacing and the overall layout of the pages are different.
Lastly, both the sizes and the aspect ratios of the images differ significantly (cf.~Table~\ref{tab:dataset_stats}) which exacerbates the challenge of different fonts and layouts even further.

\begin{table}
\renewcommand{\arraystretch}{1.3}
\centering
\caption{Statistics of the dataset images}
\begin{tabular}{l|c|c}
& Size & Aspect Ratio\\\hline
Minimum & $1000 \times 1187$ & $1 : 1.187$ \\
Median & $2495 \times 3861$ & $1 : 1.649$ \\
Maximum & $3477 \times 4945$ & $1 : 1.878$ \\

\end{tabular}
\label{tab:dataset_stats}
\end{table}

The handwritten annotations in the images are of various nature.
They are often comments on the sides and underlines of the printed text.
However, sometimes the annotations are also written between printed text lines.
Sometimes, the document may even consist of handwriting only (cf.~Fig.~\ref{fig:dataset}, which is a note written on an empty page).
Also, the indentations of the handwritten annotations change, sometimes even within the same document page.
A challenge regarding the underlines is to distinguish handwritten lines from horizontally printed lines.

\begin{figure}
    \centering
        \fbox{\includegraphics[width=0.95\linewidth]{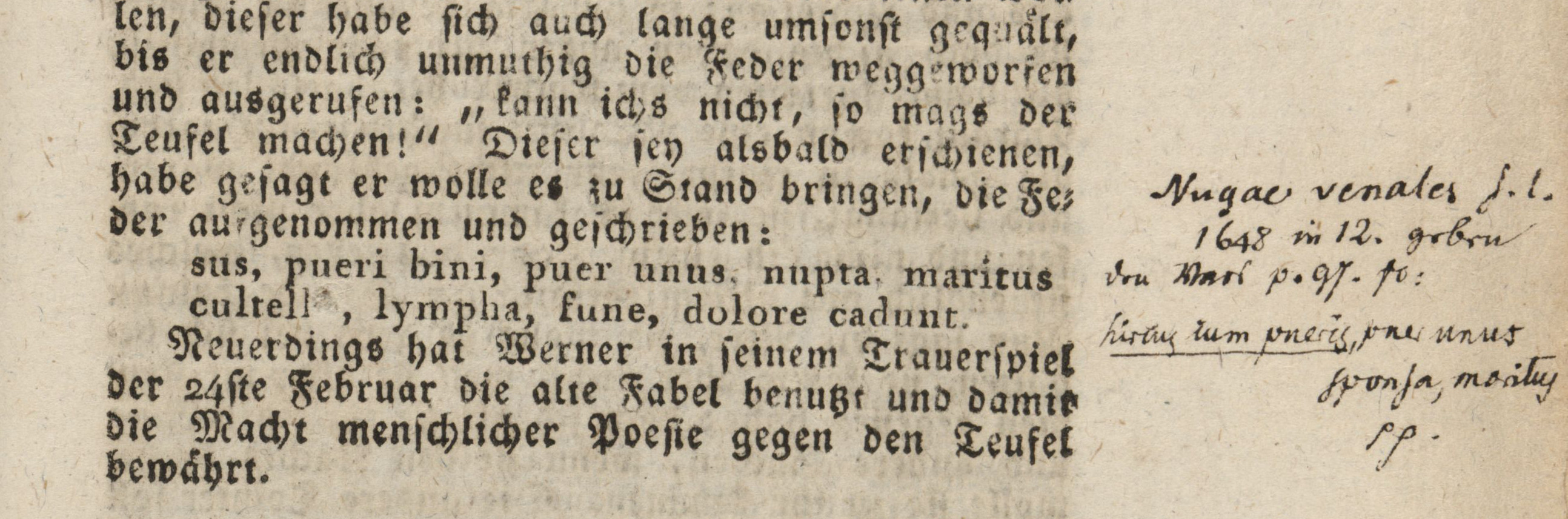}}
        \par\smallskip
        \fbox{\includegraphics[width=0.95\linewidth]{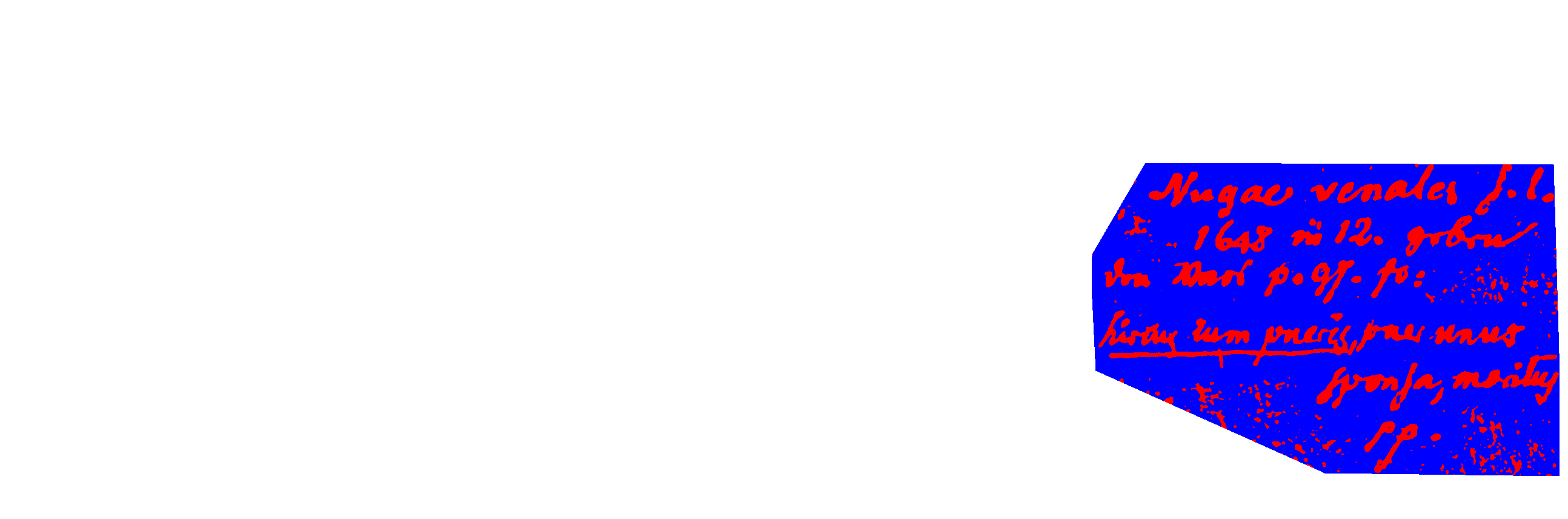}}
    \caption{Sample patch of the dataset and the color-coded \ac{gt}. Red pixels in the \ac{gt} depict annotations, blue pixels depict ambiguous regions and white pixels in the \ac{gt} are considered as background.}
    \label{fig:gt_explanation}
\end{figure}

Since the \ac{gt} is available in PAGE format, we first convert it to two-color images. To achieve this, we use the DIVA-HisDB-PixelLevelLayout\footnote{\url{https://github.com/DIVA-DIA/DIVA-HisDB-PixelLevelLayout}} tool. It binarizes the images and overlays it with the polygons which are described in the \ac{gt} PAGE file. The pixels which are black after the binarization and within the annotation polygons are considered annotation pixels, the pixels which are inside the polygons and white are marked as \emph{ambiguous} and are not considered in evaluation (cf.~Fig.~\ref{fig:gt_explanation}). All pixels outside of the \ac{gt} polygons are labeled as background. Since the binarization of the images is not perfect, the generated \ac{gt} images can contain some incorrectly labeled pixels. However, these are very few and can be neglected.

All training and testing images with their corresponding \ac{gt} annotations in both PAGE and PNG format can be downloaded at \url{http://tc11.cvc.uab.es/datasets/AnnotationDB_1}.

\section{Network Architectures}
\label{sec:networks}

In this section, we describe the \ac{fcnn} architectures we used to segment the images into the two classes: Handwritten annotations and background.

\subsection{FCN-8s}
The FCN-8s is a multi-stream \ac{fcnn} architecture that was proposed by Long \etal in 2015~\cite{Long_2015_CVPR} and scores a mean \ac{iou} of $65.4\,\%$ on the segmentation task of the Pascal \ac{voc} Challenge~\cite{everingham2015pascal}.
The architecture is based on the famous VGG-16~\cite{simonyan2014very} architecture which performed best on the localization task of the \ac{ilsvrc}~\cite{russakovsky2015imagenet} in 2014.

\begin{figure*}[t]
    \centering
        \includegraphics[width=0.95\linewidth]{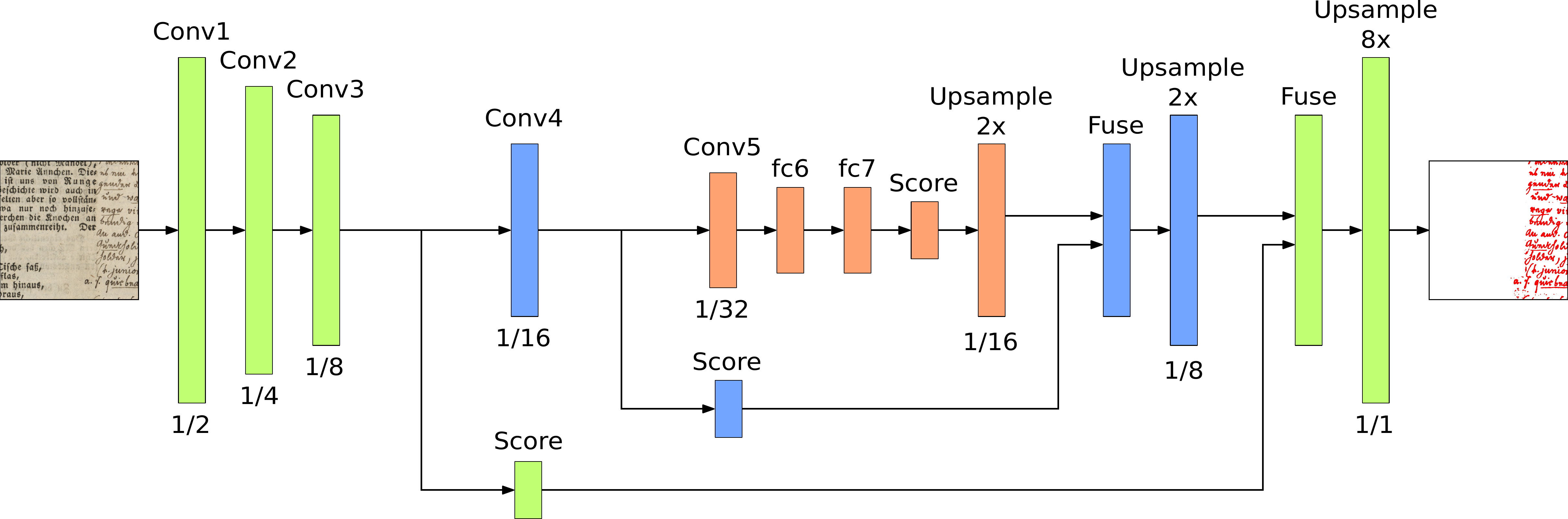}
    \caption{Segmenting an image with the multi-stream network FCN-8s. The five convolutional stacks each reduce the image size by a factor of $2$. With bilinear upsampling, the original size of the input image can be restored for dense prediction. The score layers are $1 \times 1$ Convolutions which reduce the number of feature maps to $2$, as we have $2$ classes. The output is, therefore, a two-channel image with the same size as the input image.}
    \label{fig:fcn-8s}
\end{figure*}

Just like VGG-16, FCN-8s employs five stacks of convolutional layers with appended max-pooling. Each of these pooling layers reduces the height and width of the of the feature maps by a factor of $2$. Thus, the feature maps after the fifth pooling layer are of size $(1/32 * H) \times (1/32 * W)$, with $H$ and $W$ denoting the height and width of the input image, respectively (cf.~Fig.~\ref{fig:fcn-8s}).
While VGG-16 feeds the features of the last pooling layer to a stack of three fully-connected layers for classification, the feature maps in FCN-8s are processed by a $1 \times 1$ convolutional layer which produces one feature map per class, \ie class scores. These feature maps are then upsampled with bilinear interpolation to generate pixel-wise classification results for segmentation.

Additionally to the stream described above, the network contains two more streams to incorporate lower level features into the final classification. This has proven to boost the network performance~\cite{Long_2015_CVPR,8270268}. Specifically, the network uses the features from the third and the fourth stack of convolutions and fuses them with the features of the final stack of convolutions. Since these feature maps are of a different size than the feature maps in the main stream, the fusion and upsampling happens in multiple steps (cf.~Fig.~\ref{fig:fcn-8s}). Fusion, in this case, is a simple addition of the feature maps.

\subsection{DeepLab v2}
Deeplab v2\cite{chen2018deeplab} is an \ac{fcnn} which is based on the architecture of ResNet-101\cite{DBLP:journals/corr/HeZRS15} and was proposed by Chen \etal \ in 2016. Using \emph{atrous convolution}, \emph{multi-scale image representations} and \acp{crf}~\cite{krahenbuhl2011efficient}, the method scores $79.7\,\%$ mean \ac{iou} on the segmentation task of the Pascal \ac{voc} Challenge~\cite{everingham2015pascal}. 

Atrous convolutions, also known as dilated convolutions, have a practical advantage over regular convolution and max-pooling layers, as the spatial resolution of the feature map can be retained without any upsampling involved~\cite{yu2015multi}. Although using atrous convolutions throughout the network can lead to feature maps at the original image resolution, it is computationally more expensive to keep full-sized feature maps throughout the entire network. Thus, DeepLab v2 is a hybrid network architecture that includes both atrous convolutions and standard convolutions with max-pooling and bilinear interpolation.

Chen \etal \ further propose multi-scale image processing with their network architecture. They let rescaled versions of the original image be processed by multiple branches of the network. All versions are then upsampled by bilinear interpolation to obtain the original image resolution and fused by selecting the maximum response across the different scales.

Lastly, a fully-connected \ac{crf}~\cite{krahenbuhl2011efficient} is applied as a post-processing step to further improve the results.

\subsection{\ac{pspnet}}
Like DeepLab v2, \ac{pspnet}~\cite{zhao2017pyramid} is an \ac{fcnn} architecture that is based on ResNet~\cite{DBLP:journals/corr/HeZRS15} with atrous convolutions. It was proposed in different versions, building on top of ResNet-50, ResNet-101, ResNet-152 and ResNet-269. On the Pascal \ac{voc} Challenge, the method scores $85.4\,\%$ mean \ac{iou}.
However, since the performance boost of the deeper versions of \ac{pspnet} over the ResNet-50 based version is less than $1\,\%$ (cf.~\cite{zhao2017pyramid}), we use the latter for our two-class segmentation task.

The main difference from DeepLab v2 is that \ac{pspnet}, as its name suggests, uses a pyramid pooling module to fuse features from different sub-regions of the image at different scales, while DeepLab v2 uses multi-scale image processing.

\section{Experiments}
\label{sec:experiments}

We train and evaluate the \acp{fcnn} described in Section~\ref{sec:networks} on semantic segmentation.
Thereby we explore transfer learning, binarized images, and data augmentation techniques.
All experiments are performed on an NVIDIA Titan X GPU.

We report the mean \ac{iou} score over the entire test set.
For each class, the \ac{iou} score is:
\begin{equation*}
    \frac{TP}{(FP + TP + FN)}
\end{equation*}
where $TP$, $FP$, and $FN$ denote the true positives, false positives, and false negatives, respectively.

The mean \ac{iou} is simply the average over all classes. We compute mean \ac{iou} as Long \etal~\cite{Long_2015_CVPR}:
\begin{equation*}
    \text{mean \ac{iou}} = \frac{1}{n_{cl}} * \sum_i\frac{{n_{ii}}}{(t_i + \sum_j{n_{ji}} - n_{ii})}
\end{equation*}
where $n_{cl}$ denotes the number of classes, $n_{ij}$ denotes the number of pixels of class $i$ predicted to belong to class $j$, and $t_i$ denotes the total number of pixels of class $i$.

The evaluation is performed using the DIVA Layout Analysis Evaluator tool~\cite{2017arXiv171201656A}. This tool is used for document segmentation for multi-labeled pixel \ac{gt} and has also been used for the ICDAR2017 Layout Analysis for Challenging Medieval Manuscripts~\cite{simistira2017icdar2017}.
It appropriately fits our requirements as it takes care of ambiguous regions (cf.~Fig.~\ref{fig:gt_explanation}).

\subsection{Transfer Learning and Finetuning}
As our dataset contains only $40$ images for training, it is natural to exploit other datasets for pretraining and then finetune the network parameters on our target dataset.
For pretraining, we exploit both, the \ac{ilsvrc} dataset which contains $1.2$ million real-world images~\cite{russakovsky2015imagenet} and the DIVA-HisDB dataset which contains $120$ images from historic documents~\cite{simistira2016diva}.

When initializing the weights with \ac{ilsvrc} pretraining, the convolutional layers just keep their weight matrices.
However, since the original network architectures use fully-connected layers for classification and our networks are fully convolutional, we have to convert the fully connected layers to convolutional layers to benefit from the pretraining.
The trick is to interpret these layers as convolutional layers with kernels that cover the entire input region (cf.~\cite{Long_2015_CVPR}). Now, all we have to do is to reshape the corresponding weight matrices and we have a \ac{fcnn}.

\subsection{Data Augmentation}
To compensate for our limited training data, we use multiple data augmentation techniques to artificially increase our training data.

First, we use simple random cropping.
Since the networks are too large to fit in the GPU memory with large images (cf.~Section~\ref{sec:dataset}), we crop patches of $512 \times 512$ pixels from the images.

Second, we employ the data augmentation technique that was proposed by Szegedy \etal \ to train the Inception networks~\cite{Szegedy_2015_CVPR}.
This is a more sophisticated data augmentation technique which also adds some scaling and aspect ratio invariance to the images.
We found these properties particularly useful for our heterogeneous dataset (cf.~Section~\ref{sec:dataset}).
The method samples patches from the images uniformly at random with the following limitations.
The area of the patch covers $8\,\%$ to $100\,\%$ of the whole image and its aspect ratio is between $\frac{3}{4}$ and $\frac{4}{3}$. This patch is then scaled to $512 \times 512$ pixels. 

In a last set of experiments, we binarize the training images, as this has been used frequently for segmentation tasks where the background color is more or less uniform~\cite{Zhong:1995:LTC:844379.844664}. The images are binarized using adaptive thresholding and again cropped to $512 \times 512$ pixels to fit in the GPU memory. We manually verify the representativeness of the samples and the areas of interest in the images.

\begin{table}
\renewcommand{\arraystretch}{1.3}
\centering
\caption{Performance of the networks on the new dataset with different pretraining and data augmentation.}
\begin{tabular}{l|c|c|c}
 & Pretraining & Data Augmentation & Mean \ac{iou}\\\hline
FCN-8s & \ac{ilsvrc} & Binarization & $63.0\,\%$ \\\hline
FCN-8s & DIVA-HisDB & Random Cropping & $70.0\,\%$ \\\hline
\ac{pspnet} & \ac{ilsvrc} & Random Cropping & $87.7\,\%$ \\\hline
DeepLab v2 & \ac{ilsvrc} & Random Cropping & $89.2\,\%$ \\\hline
FCN-8s & \ac{ilsvrc} & Random Cropping & $91.3\,\%$ \\\hline
FCN-8s & \ac{ilsvrc} & Inception & $\mathbf{95.6}\,\boldsymbol{\%}$
\end{tabular}
\label{tab:results}
\end{table}

\section{Results}
\label{sec:results}
In the following, we report the results of the experiments and attempt to give reasons for the outcomes.

For evaluation of the trained networks, we pass patches of $512 \times 512$ pixels from the test images through the network in a sliding-window manner with an overlap of $128$ pixels. Predictions of pixels which are present in multiple patches are averaged.
The scores achieved by the different networks are presented in Table~\ref{tab:results}.

Table~\ref{tab:results} reveals that neither binarization of the images nor pretraining on DIVA-HisDB yields good results.
In the case of training and testing on binary images, lots of pixel information is already lost during preprocessing.
While a colored pixel can have $2^{24}$ different values, it can only be black or white in binary images.
Binarization may be helpful in cases where all the data is very similar, but for our diverse images, it is crucial to use colored images.
The diversity of our dataset could also explain, why pretraining on DIVA-HisDB resulted in a poor mean \ac{iou} of only $70\,\%$.
When trained on DIVA-HisDB, FCN-8s reaches a mean \ac{iou} of more than $90\,\%$ on the DIVA-HisDB test set, which is comparable to the \sota.
However, if these weights are used as initialization for training on our dataset, the network is unable to generalize over the new document types, because it is too inclined to the homogeneous DIVA-HisDB images.

\begin{figure}
    \centering
        \fbox{\includegraphics[width=0.94\linewidth]{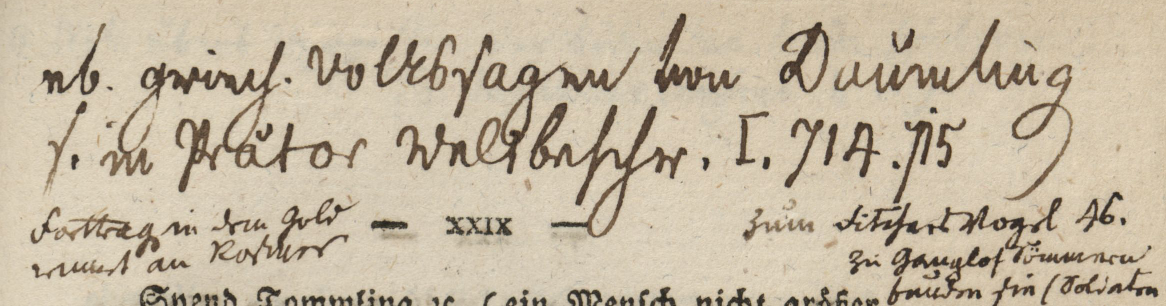}}
        \par\smallskip
        \fbox{\includegraphics[width=0.94\linewidth]{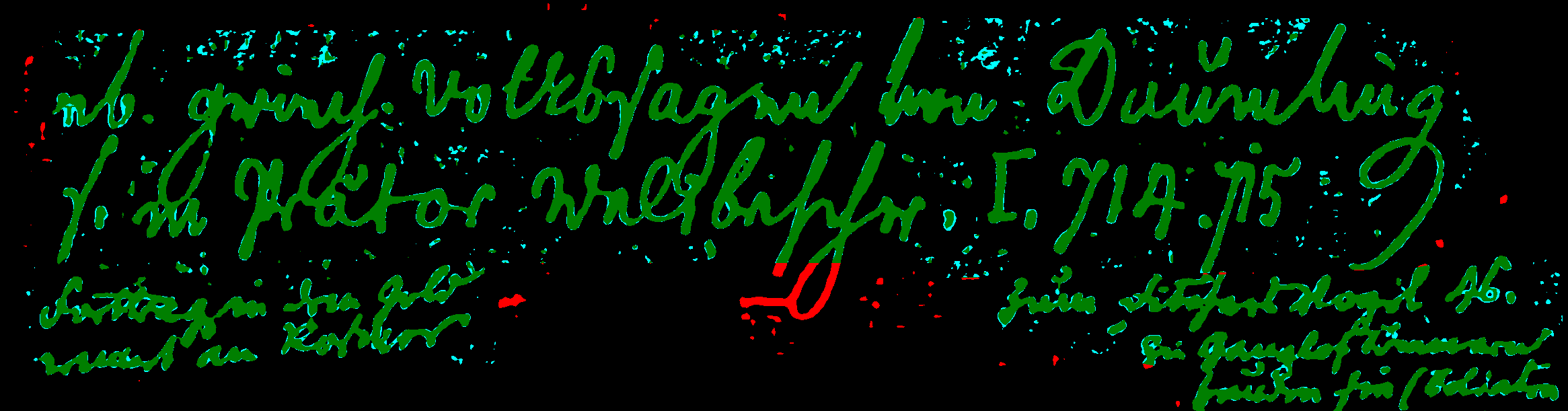}}
    \caption{Sample from the test set (top) and visualization of the prediction of our best model (bottom). Green pixels are true positives, black pixels are true negatives, red pixels are false positives and blue pixels are false negatives.}
    \label{fig:results}
\end{figure}

Pretraining the networks on the \ac{ilsvrc} dataset yields overall good results, as the trained convolutional filters are very robust and generalize well thanks to the amount of training data.

\section{Discussion}
\label{sec:discussion}
The results of the work signify that the current segmentation approaches and data augmentation techniques work well with the proposed complex dataset. The dataset contains images from multiple sources and a variety of different types of artifacts.

Another strong contribution of the paper is the introduction of the new dataset.
While it may seem that the number of pages is small, the task is to label every pixel and hence the dataset is large enough.
Also the results show that with the reasonable amount of augmentation we are able to achieve great results. 

Furthermore, it is worth mentioning the noise in the \ac{gt}. The \ac{gt} generation is well explained in section~\ref{sec:dataset}.
Although there exists some noise in the \ac{gt}-pixels due to the binarization, it does not affect the evaluation significantly. 
Furthermore, the main purpose of the proposed work is to signify how well we can perform on a difficult dataset of historical documents.
However, the improvement of the \ac{gt} can be done in the future with the provided dataset, as discussed in section~\ref{sec:conclusion}.


As far as the results are concerned, the \ac{fcnn} based approaches are well capable of segmenting the documents.
While DeepLab v2 and \ac{pspnet} perform significantly better on real-world images than FCN-8s (cf.~Section~\ref{sec:networks}), FCN-8s performs best on our dataset.
This result is consistent with a finding by Afzal \etal~\cite{afzal2017cutting} who found VGG-16 better suited for document image classification, then ResNet-50.
When analyzing the errors visually, we find that the ResNet based networks have problems detecting fine pencil annotations.
This relates to one of the failure modes discovered by Chen \etal, where their best performing model ``fails to capture the delicate boundaries of objects, such as bicycle and chair''~\cite{chen2018deeplab}.

With a data augmentation technique that fits well to our dataset, we are able to reduce the error by half and achieve a final mean \ac{iou} score of $95.6\,\%$.
At visual inspection, our best model is almost perfect in detecting the handwritten annotations. Most errors come from noisy or incorrect \ac{gt} (cf.~Fig.~\ref{fig:results}).

Our results on this difficult dataset is one step forward towards achieving a general purpose segmentation method that could work with most of the documents. 


\section{Conclusion and Future Work}
\label{sec:conclusion}

In this work, we have presented a new and challenging dataset for document layout analysis in terms of semantic segmentation.
Furthermore, we have trained and evaluated multiple combinations of different network architectures, weight initializations, and data augmentation strategies.
The models trained in this paper do not include any domain-specific post-processing methods.
Employing domain-specific knowledge would further improve the results, as has been shown in~\cite{simistira2017icdar2017} recently.

An important direction for future work is the analysis of the generalization and scalability of our methods.
While our dataset is already of diverse nature, it would be interesting to see if it also applies to documents of other libraries and countries.
Furthermore, specialized methods could be adopted from our classifier to work on specific types of books and to incrementally work better after a few pages.

As it is not feasible to get correct binarization for all the images in the dataset with automatic approaches, a possible improvement in the \ac{gt} pixels could be performed by manually cleaning the binarized documents.

Finally, it would be good to integrate our approach into production systems.
A possible scenario is that human experts can inspect all annotations found by the system, slightly correct them (enabling active learning and human-in-the-loop systems), and integrate the findings into higher-level research questions in the humanities, such as the change of annotation-behavior over the centuries.

\section*{Acknowledgments}
We thank Insiders Technologies GmbH for providing the hardware to conduct the experiments described in this paper.
We also want to thank Abhash Sinha and Shridhar Kumar for performing the experiments with \ac{pspnet}.
Finally, we thank the following libraries for providing us with the digitized images for the database: Universit\"atsbibliothek der Humboldt-Universit\"at zu Berlin; Universit\"atsbibliothek Kassel; Staatsbibliothek Berlin; and Staatsarchiv Marburg.

\bibliographystyle{IEEEtran}
\bibliography{literature.bib}

\end{document}